\documentclass{article}

\usepackage{arxiv}

\usepackage[utf8]{inputenc} 
\usepackage[T1]{fontenc}    
\usepackage{hyperref}       
\usepackage{url}            
\usepackage{booktabs}       
\usepackage{amsfonts}       
\usepackage{nicefrac}       
\usepackage{microtype}      
\usepackage{amsmath} 
\usepackage{cleveref}       
\usepackage{graphicx}
\usepackage{natbib}
\usepackage{doi}
\usepackage{tabularx}
\usepackage{ragged2e} 
\usepackage{array}
\usepackage{float} 

\newcolumntype{C}{>{\Centering\arraybackslash}X}

\newcommand{\ArticleTitle}{Game-Theoretic Gradient Control for Robust Neural Network Training}

\newcommand{\AMZ}{Maria Zaitseva}

\newcommand{\AIT}{Ivan Tomilov}

\newcommand{\ANG}{Natalia Gusarova}

\newcommand{\DatasetsCount}{Ten}
\newcommand{\datasetscount}{ten}

\title{\ArticleTitle}

\newif\ifuniqueAffiliation
\uniqueAffiliationtrue

\ifuniqueAffiliation 
\author{ \href{https://orcid.org/0009-0008-3063-3331}{\includegraphics[scale=0.06]{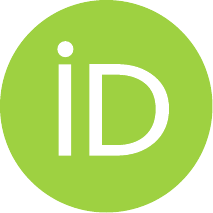}\hspace{1mm}\AMZ} \\ 
	Faculty of Applied Informatics \\
	ITMO University \\
	\texttt{tail-call@yandex.ru} \\
	\And
	\href{https://orcid.org/0000-0003-1886-2867}{\includegraphics[scale=0.06]{orcid.pdf}\hspace{1mm}\AIT} \\
	Faculty of Applied Informatics \\
	ITMO University \\
	\texttt{ivan-tomilov3@yandex.ru} \\
	\AND
	\href{https://orcid.org/0000-0002-1361-6037}{\includegraphics[scale=0.06]{orcid.pdf}\hspace{1mm}\ANG} \\
	Faculty of Applied Informatics \\
	ITMO University \\
	\texttt{nfgusarova@itmo.ru} \\
}
\else
\usepackage{authblk}

\newbox{\orcid}\sbox{\orcid}{\includegraphics[scale=0.06]{orcid.pdf}} 
\author[1]{%
	\href{https://orcid.org/0009-0008-3063-3331}{\usebox{\orcid}\hspace{1mm}\AMZ\thanks{\texttt{hippo@cs.cranberry-lemon.edu}}}%
}
\author[1,2]{%
	\href{https://orcid.org/0000-0000-0000-0000}{\usebox{\orcid}\hspace{1mm}Elias D.~Striatum\thanks{\texttt{stariate@ee.mount-sheikh.edu}}}%
}
\affil[1]{Department of Computer Science, Cranberry-Lemon University, Pittsburgh, PA 15213}
\affil[2]{Department of Electrical Engineering, Mount-Sheikh University, Santa Narimana, Levand}
\fi

\renewcommand{\undertitle}{A Preprint}

\hypersetup{
pdftitle={\ArticleTitle},
pdfsubject={cs.LG	},
pdfauthor={\AMZ, \AIT, \ANG},
pdfkeywords={Feed-forward neural networks, Open games, Noise robustness},
}

\begin{document}
\maketitle

\begin{abstract}
Feed-forward neural networks (FFNNs) are vulnerable to input noise, reducing prediction performance. Existing regularization methods like dropout often alter network architecture or overlook neuron interactions. This study aims to enhance FFNN noise robustness by modifying backpropagation, interpreted as a multi-agent game, and exploring controlled target variable noising. Our ``gradient dropout'' selectively nullifies hidden layer neuron gradients with probability \(1-p\) during backpropagation, while keeping forward passes active. This is framed within compositional game theory. Additionally, target variables were perturbed with white noise or stable distributions. Experiments on \datasetscount diverse tabular datasets show varying impacts: improvement or diminishing of robustness and accuracy, depending on dataset and hyperparameters. Notably, on regression tasks, gradient dropout (\(p=0.9\)) combined with stable distribution target noising significantly increased input noise robustness, evidenced by flatter MSE curves and more stable SMAPE values. These results highlight the method's potential, underscore the critical role of adaptive parameter tuning, and open new avenues for analyzing neural networks as complex adaptive systems exhibiting emergent behavior within a game-theoretic framework.
\end{abstract}


\section{Introduction}

Feed-forward neural networks (FFNNs) are widely employed for diverse computational tasks, yet their performance and reliability are often compromised by their inherent vulnerability to noise present in input data. To mitigate this, various regularization techniques have been developed, notably dropout, which stochastically deactivates neurons during both forward and backward passes of training \citep{Srivastava2014Dropout}. However, classical dropout inherently alters the network's architecture at each training step, potentially reducing overall accuracy by creating a sparse sub-network. Furthermore, existing regularization methods often do not explicitly account for the intricate interactions among individual neurons, which can be viewed as agents contributing to a collective outcome.

This study proposes an approach to enhance the noise robustness of FFNNs by modifying the error backpropagation process. For this, we interpret network training as a multi-agent game, an approach that allows for more nuanced control over the contribution of individual neuron-agents to error correction. This interpretation is grounded in Compositional Game Theory (CGT) \citep{cgt}, a categorical framework that enables the application of game theory in domains where systems are composed from well-defined sub-systems. Additionally, we investigate the effect of controlled noising of the target variable to explore its impact on training dynamics.

The increasing reliance on neural networks, coupled with the persistent challenge of their interpretability and understanding of their internal mechanisms, underscores the relevance of this research. Despite their remarkable successes, neural networks frequently function as ``black boxes,'' impeding optimization, diagnosis, and reliability enhancement. Traditional analysis methods typically focus on isolated aspects, failing to offer a holistic view of network operation. By modelling neural network elements as agents in an interactive game, CGT presents a promising avenue to overcome these limitations. Decomposing neural architectures into sub-systems that can be analyzed independently or as a part of a whole requires a framework for meaningfully composing them. We demonstrate how~FFNNs can be constructed in terms of open games, the building block of CGT, and~discuss implications following from their structure.

Our approach, particularly the concept of gradient dropout, builds upon and is distinguished from prior work in several key aspects. A game-theoretic interpretation of standard dropout was proposed by \citet{zhang2021dropout}, who demonstrated that dropout suppresses interactions between features, linking this directly to a reduction in overfitting. Their work, however, focuses on interpreting standard dropout, whereas our approach modifies the backpropagation process itself to specifically enhance noise robustness by managing neuron-agent contributions.

While various methods exist under the term ``gradient dropout,'' our technique is distinct. For instance, one method applies gradient dropout in the context of meta-learning to improve few-shot adaptation \citep{tseng2020gradient}. Our method, in contrast, focuses on enhancing the noise robustness of standard FFNNs in a single-task supervised learning setting by modifying the primary backpropagation process, not a specialized meta-learning adaptation step. Another approach, adaptive trainable gradient dropout, uses auxiliary networks to learn masks that induce sparsity by freezing specific connections \citep{avgerinos2023adaptive}. Our method uses a fixed, stochastic regularization by randomly nullifying gradients for entire neurons, interpreted as randomized rewards for agents, rather than learning a deterministic sparsity mask. Finally, GradDrop was designed for multi-task learning, where it resolves gradient conflicts by masking gradients based on their sign \citep{zhao2020graddrop}. Our method is designed for single-task learning and randomly nullifies the entire gradient for a selected neuron, irrespective of sign. In summary, our method is unique in its application of stochastic, backward-pass-only gradient nullification to enhance noise robustness, all interpreted through the lens of randomized rewards within compositional game theory.

While game-theoretic approaches have been applied to various machine learning domains, such applications are often fragmented and limited to specific problems. Therefore, the development of a unified approach for analyzing neural networks based on compositional game theory is highly relevant and offers significant potential to advance machine learning methodologies by moving towards more interpretable and robust AI systems.

\section{Materials and Methods}

This section details the experimental design, methodology, and theoretical foundations of the proposed modifications to FFNNs. The objective of this study is to systematically investigate the relationship between input noise, controlled perturbations of the backpropagation mechanism, and stochastic modifications of the target variable on the predictive accuracy and robustness of FFNNs in both classification and regression tasks.

\subsection{Neural Network Architecture}
A fully connected neural network (FFNN) architecture was employed for all experiments, consisting of three layers:
\begin{enumerate}
    \item \textbf{Input Layer:} Dimensionality determined by dataset features.
    \item \textbf{Hidden Layers:} 5 hidden layers, each with $N_h = 150$ neurons and with Rectified~Linear Unit (ReLU) activation:
    
\begin{equation}
    \label{eq:relu}
    \phi(z) = \max(0, z)
\end{equation}

    \item \textbf{Output Layer:} Single neuron for regression; number of classes for classification.
\end{enumerate}

\subsection{Controlled Backpropagation Modification: Gradient Dropout}
Here we introduce Gradient Dropout, a core modification to the backward pass that stochastically gates gradients in hidden and output layers, inspired by compositional game theory (CGT) and parametric lenses \citep{dlpl_cruttwell24}. We will show it is possible to interpret parametric lenses as a concrete implementation of abstract open games.

\subsubsection{Formalization with Parametric Lenses and Open Games}

Neural network are usually seen as monolithically optimized entities. In this paper we adopt a more granular, agent-based perspective, viewing a neural network as a sequential game played by its layers. During one training iteration, each layer, acting as a distinct player, performs:

\begin{itemize}
	\item forward pass -- observes the output of the previous layer, makes a strategic choice;
	\item backward pass -- receives a utility signal (a gradient) from the subsequent layer, to which it provides a ``best response'' by adjusting its own parameters.
\end{itemize}
	
We formally model each dense layer $D_i$ of the network as a distinct open game $G_i: (X_i, S_i) \to (Y_i, R_i)$, where the full network is their sequential composition $G_{NN} = G_n \circ \dots \circ G_1$. The components of a single game $G_i$ (representing layer $D_i$) are defined as follows:
\begin{itemize}
    \item \textbf{Strategy Space ($\Sigma_{G_i}$):} The layer's strategy is its set of weights and biases. A specific strategy $\sigma_i = (W^{(i)}, b^{(i)})$ is a choice of these parameters.
    \begin{equation}
        \Sigma_{G_i} = \mathbb{R}^{N_i \times N_{i-1}} \times \mathbb{R}^{N_i}
    \end{equation}
    \item \textbf{Observation ($X_i$) and Choice ($Y_i$):} The layer observes the input activations $a^{(i-1)}$ from the previous layer and chooses to produce output activations $a^{(i)}$ for the next layer.
    \begin{equation}
        X_i = \mathbb{R}^{N_{i-1}}, \quad Y_i = \mathbb{R}^{N_i}
    \end{equation}
    \item \textbf{Play Function ($P_{G_i}: \Sigma_{G_i} \times X_i \to Y_i$):} This is the layer's forward pass, mapping a strategy and an observation to a choice.
    \begin{equation}
        P_{G_i}((W^{(i)}, b^{(i)}), a^{(i-1)}) = a^{(i)} = \phi(W^{(i)} a^{(i-1)} + b^{(i)})
    \end{equation}
    where $\phi$ is an activation function like ReLU \ref{eq:relu}.
    \item \textbf{Utility ($R_i$) and Coutility ($S_i$):} In backpropagation, utility is the backward-flowing gradient of the loss with respect to the layer's output, while coutility is the gradient of the loss with respect to its input, which serves as the utility for the preceding player.
    \begin{equation}
        R_i = (\mathbb{R}^{N_i})', \quad S_i = (\mathbb{R}^{N_{i-1}})' \quad (\text{The spaces of output and input gradients})
    \end{equation}
    \item \textbf{Coplay Function ($C_{G_i}: \Sigma_{G_i} \times X_i \times R_i \to S_i$):} This computes the coutility from the strategy, observation, and incoming utility. It is the part of the backward pass that computes the input gradient.
    \begin{equation}
        C_{G_i}((W^{(i)}, b^{(i)}), a^{(i-1)}, \nabla_{a^{(i)}}L) = \nabla_{a^{(i-1)}}L
    \end{equation}
    \item \textbf{Best Response Function ($B_{G_i}$):} This defines how the layer updates its strategy to improve its outcome, which is precisely gradient descent (or a variant like Adam). The best response $\sigma'_i$ is the updated set of parameters.
    \begin{equation}
        \sigma'_i \in B_{G_i}(a^{(i-1)}, k) \quad \text{iff} \quad \sigma'_i = \sigma_i - \eta \cdot \nabla_{\sigma_i} L
    \end{equation}
    where $k: Y_i \to R_i$ is the continuation representing the utility landscape, and $\nabla_{\sigma_i} L$ is the parameter gradient.
\end{itemize}
This game-theoretic interpretation is enabled by the formal mechanics of parametric lenses \citep{dlpl_cruttwell24}, where the \texttt{get} map of a lens corresponds to the play function, and the \texttt{put} map computes both the parameter gradients for the best response and the input gradients for the coplay function.

Now we can precisely describe our method. A standard fully connected layer $k$ computes its pre-activation $z^{(k)}$ and activation $a^{(k)}$ as:
\begin{equation}
    z^{(k)} = W^{(k)} a^{(k-1)} + b^{(k)}
\end{equation}
\begin{equation}
    a^{(k)} = \phi(z^{(k)}) = \text{ReLU}(z^{(k)})
\end{equation}
The standard backward pass propagates the gradient of the loss function $L$. For each neuron $j$ in layer $k$, the local gradient (error) $\delta_j^{(k)}$ is:
\begin{equation}
    \delta_j^{(k)} = \frac{\partial L}{\partial z_j^{(k)}} = \frac{\partial L}{\partial a_j^{(k)}} \phi'(z_j^{(k)})
\end{equation}
where $\phi'(z_j^{(k)}) = 1$ if $z_j^{(k)} > 0$ and $0$ otherwise for ReLU. The gradient for a weight $w_{ji}^{(k)}$ is:
\begin{equation}
    \frac{\partial L}{\partial w_{ji}^{(k)}} = \delta_j^{(k)} a_i^{(k-1)}
\end{equation}
Our proposed Gradient Dropout method modifies this gradient calculation by introducing a stochastic multiplier $m_j \in \{0, 1\}$ for each neuron $j$ in the hidden and output layers, drawn from a Bernoulli distribution with hyperparameter $p$:
\begin{equation}
    m_j \sim \text{Bernoulli}(p)
\end{equation}

\begin{figure}[htbp]
\centering
\includegraphics[width=10.0 cm]{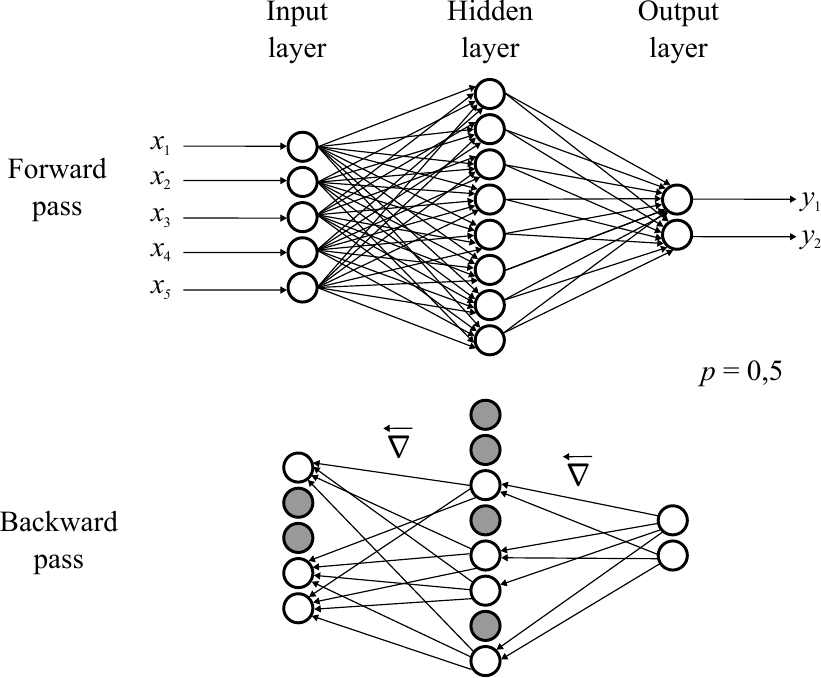}
\caption{Illustration of gradient dropout in a feed-forward neural network. The network consists of one hidden layer, five inputs ($x_1, \ldots, x_5$), and two outputs ($y_1, y_2$). During the backward pass, each unit deactivates with a probability of $p = 0.5$; units remain active during the forward pass. $\overleftarrow{\nabla}$~represents the back-propagated gradient. \label{fig1}}
\end{figure}

Figure~\ref{fig1} illustrates Gradient Dropout.
 
The modified gradient used for updating the weight $w_{ji}^{(k)}$ becomes:
\begin{equation}
    \label{eq:modified_gradient}
    \left(\frac{\partial L}{\partial w_{ji}^{(k)}}\right)_{\text{modified}} = m_j \cdot \delta_j^{(k)} a_i^{(k-1)}
\end{equation}

This operation is implemented within a custom \texttt{torch.autograd.Function}. The forward pass remains identical, but the backward pass multiplies the incoming gradient for each neuron by its corresponding mask $m_j$. In the game-theoretic context, this effectively stochasticizes the communication between agents by randomly nullifying components of the utility and coutility signals passed between layers. The effective gradient for an entire layer's weights $W^{(k)}$ can be conceptualized as:
\begin{equation}
    (\nabla_{W^{(k)}} L)_{\text{modified}} = M^{(k)} * \nabla_{W^{(k)}} L_{\text{original}}
\end{equation}
where $M^{(k)}$ is a matrix where each row $j$ is scaled by $m_j = M^{(k)}_{ji}$, and $*$ denotes element-wise multiplication.

The hyperparameter $p \in \{0, 0.01, 0.05, 0.5, 0.9, 0.95, 0.99\}$ was systematically varied to investigate its impact on noise robustness. For $p=1$, the method degenerates to standard backpropagation; for $p=0$, all gradients for the affected layers are zeroed and no learning happens.

\subsubsection{Game-Theoretic Analysis: Learning Composable Primitives versus Statistical Shortcuts}
To provide a theoretical motivation for our approach, we can model the interaction between neurons as a multi-agent game. This allows us to analyze how different regularization schemes affect the learning dynamics. We reframe the Public Goods Game to model a fundamental trade-off in representation learning: the choice between learning disentangled, composable primitives versus entangled, statistical shortcuts. This challenge, where learning signals for compositional and non-compositional features become coupled in standard architectures, has been formally analyzed by \cite{jarvis2023specialization}.

\begin{itemize}
    \item \textbf{Players}: The $N$ neurons in a layer $k$.
    \item \textbf{Actions}: Each neuron $i$ effectively chooses a strategy $a_i \in \{C, D\}$:
    \begin{itemize}
        \item \textbf{Strategy C (Compose)}: The neuron specializes in learning a single, disentangled, and composable feature (e.g., ``redness,'' ``circularity''). This is a high-risk, high-reward strategy. The feature may only be useful for the final task when combined with other primitives learned by a ``coalition'' of other neurons. However, if successful, this is the only path to true \textbf{systematic generalization}.
        \item \textbf{Strategy D (Detour)}: The neuron learns a statistically frequent but entangled feature that serves as a shortcut (e.g., if most red objects in the training set are cars, it learns a ``red-car-blob'' feature). This is a low-risk, low-reward ``generalist'' strategy. It is ``general'' only in its applicability \textit{across the training set}, where this spurious correlation provides a safe and easy way to reduce loss. This learned feature is, by its nature, \textbf{non-compositional} and will fail to generalize systematically. That is, it will not help identify a~``red~ball''.
    \end{itemize}

    \item \textbf{Payoff Function}: The payoff $u_i$ is the marginal reduction in the global loss function attributable to neuron $i$. The formal structure of the Public Goods Game is preserved, but its components are now precisely defined:
\begin{equation}
    u_i(a_i, \boldsymbol{a}_{-i}) = \frac{\kappa}{N} \sum_{j=1}^{N} v(a_j) - c(a_i)
\end{equation}
    where $v(a_j)=1$ if neuron $j$ chooses to Compose ($a_j=C$), and $v(a_j)=0$ if it chooses to Detour.
    \begin{itemize}
        \item $\kappa$ -- benefit factor. It is the large potential loss reduction from a successful coalition that learns a complete set of composable primitives.
        \item $c$  -- compose cost. It represents the \textbf{Risk of Specialization}. A neuron choosing to Compose (e.g., learn ``redness'') risks making zero contribution to the final output if other neurons fail to learn the complementary primitives (``car,'' ``ball,'' ``square''). The Detour strategy has zero risk $c(D)=0$ because it is tied to a known, frequent pattern in the training data.
    \end{itemize}
\end{itemize}

This framework reveals the game-theoretic roots of the ``coupling'' problem. A dense network, unable to segregate the payoffs from these two strategies, defaults to the lower-risk Detour strategy, thus failing to achieve systematicity. The condition for the ``Tragedy of the Commons'' $c > \kappa/N$ now has a clear interpretation: the \textbf{risk of failed co-adaptation outweighs the neuron's individual share of the potential collective reward}, driving the system towards learning non-compositional shortcuts.

\textbf{Case 1: Standard Backpropagation and the Detour Nash Equilibrium}

In a standard training setting, analogous to a single-shot game, each neuron (player) acts to maximize its immediate, individual payoff—the marginal reduction in global loss. A rational neuron will thus opt for the \textbf{Detour (D)} strategy, because its individual cost of contributing to a composable primitive ($c$, the compose cost) is, under typical conditions for systematicity challenges, greater than the individual share of the collective benefit $\kappa/N$ it helps create. This situation, where $c > \kappa/N$, means the risk of specializing and failing to co-adapt outweighs the individual neuron's share of the potential global reward.

Since this rational logic applies to all neurons in isolation, the strategy profile where all neurons choose the Detour strategy, $(\boldsymbol{a} = D, D, \dots, D)$, constitutes the unique and dominant-strategy Nash Equilibrium. This equilibrium corresponds to the network settling into a poor local minimum in the loss landscape. All neurons adopt statistical shortcuts, leading to minimal true compositional learning and failure to achieve systematic generalization. This phenomenon aligns with the "coupling" problem described by \cite{jarvis2023specialization}, where the network defaults to learning entangled features that are effective only on the training set.

\textbf{Case 2: The Effect of Altering Player Presence via Classical Dropout}

Classical dropout, which stochastically deactivates neurons during the forward pass with probability $1-p_d$, is equivalent to forcing a player into the \textbf{Detour strategy} (as they make no contribution to composable primitives). For a neuron considering the \textbf{Compose strategy}, the expected number of other neurons also choosing Compose is now reduced. This makes the expected payoff from the Compose strategy even lower, as it significantly increases the risk of unreciprocated specialization (that is, paying $c$ without sufficient complementary primitives being present).

Consequently, classical dropout powerfully reinforces the non-compositional $(\boldsymbol{a} = D, D, \dots, D)$ equilibrium. Its primary benefit is not in enabling the discovery of more optimal, compositional equilibria, but rather in preventing the formation of fragile, co-adapted "coalitions" of neurons that might learn spurious, entangled features specific to the training data (overfitting). It forces neurons to learn features that are robust and useful in isolation, even when other potential "partners" are absent, thereby improving generalization to slight perturbations in input data.

\textbf{Case 3: The Effect of Altering Payoff Information via Gradient Dropout}

Gradient Dropout proposed in this fundamentally changes the game's dynamics. All neurons are always present and contribute during the forward pass. However, the payoff signal (the gradient) is received by each neuron $i$ only with probability $p$.

The expected payoff for neuron $i$ becomes:

$$E[u_i] = p \cdot u_i$$

In a single-shot game, this multiplication by a constant does not change the order of preferences, and the Nash Equilibrium remains $(\boldsymbol{a} = D, D, \dots, D)$. However, neural network training is an iterated game. The ``payoff'' is the gradient, which is the information used to update the strategy for the next round.

By stochastically nullifying this information with probability $1-p$, we achieve an effect of the temporary suspension of rational adjustment.

A neuron that experimented with a Compose strategy might not immediately receive the negative feedback signal that would normally compel it to revert to Detour. This ``freezing'' of its strategy provides a window of opportunity for other neurons to also switch to Compose. If a critical mass of neurons switches during these stochastic windows, the collective benefit $\frac{\kappa}{N} \sum v(a_j)$ can grow to a point where Compose becomes the rational choice for everyone.

In this way, Gradient Dropout destabilizes inefficient Nash equilibria. It acts as an annealing mechanism, introducing noise into the learning signals themselves, allowing the system of agents to escape the basin of attraction of a poor local minimum and explore the strategy space to find more globally optimal, composable solutions. This aligns with our empirical findings where Gradient Dropout, particularly at high $p$ values, leads to more robust models, especially on complex regression tasks.

\subsection{Multi-Layer Public Goods Game Simulation}

To gain deeper insights into the emergent behaviors arising from the proposed gradient dropout and to visualize the game-theoretic dynamics discussed in the preceding section, we developed a simplified multi-agent simulation. This simulation models the collective decision-making process of neurons (agents) within a single layer, reframing the challenge of learning compositional features versus statistical shortcuts as an iterated Public Goods Game. While a simplification of the full neural network training process, this model provides an intuitive understanding of how different regularization strategies might influence the collective intelligence and robustness of the system.

In this simulation, we consider a layer composed of $N$ neurons, each acting as an individual player. Each neuron $i$ makes a strategic choice between two actions at each time step (analogous to a training iteration):
\begin{itemize}
    \item \textbf{Compose (C)}: Analogous to specializing in a single, disentangled, and composable feature. This strategy carries a cost $c$ (the `compose\_cost`) representing the risk and effort of specialization, but contributes to a collective benefit $\kappa$ (the `benefit\_factor`) if a sufficient number of other neurons also choose to compose.
    \item \textbf{Detour (D)}: Analogous to learning a statistically frequent but entangled feature (a "shortcut"). This strategy has zero individual cost but contributes less to the overall systematic generalization capacity.
\end{itemize}

The individual payoff $u_i$ for neuron $i$ is structured akin to a Public Goods Game, reflecting the marginal reduction in a global loss function:
\begin{equation}
    u_i(a_i, \boldsymbol{a}_{-i}) = \frac{\kappa \cdot k}{N} - \text{cost}(a_i)
\end{equation}
where $k$ is the number of agents choosing `Compose`, and $\text{cost}(a_i) = c$ if $a_i=C$ and $0$ if $a_i=D$.

The choice between strategies (Compose or Detour) is made stochastically using a Softmax function, influenced by the utility ($Q$) of each action and an `exploration\_temp` parameter $\tau$.

\subsubsection{Mechanics of Temperature (`exploration\_temp`) in Softmax Choice}

The temperature $\tau$ (implemented as `exploration\_temp` in our simulation) is a critical hyperparameter that controls the level of stochasticity in an agent's strategy selection based on the estimated utility of each action. The probabilities of choosing action $a_i$ (Compose or Detour) are computed via Softmax:

$$
P(a_i) = \frac{e^{Q(a_i)/\tau}}{\sum_{j} e^{Q(a_j)/\tau}}
$$

where $Q(a_i)$ is the utility of action $a_i$. This mechanism allows us to model the balance between exploration and exploitation in the agent's learning process:
\begin{itemize}
    \item \textbf{High Temperature ($\tau \gg 0$)}: As $\tau$ approaches infinity, the probabilities of all actions become nearly equal, regardless of their utilities. This promotes wide exploration of the strategy space, preventing agents from getting stuck in local optima.
    $$
    \lim_{\tau \to \infty} P(a_i) = \frac{1}{n}.
    $$
    \item \textbf{Low Temperature ($\tau \approx 0$)}: As $\tau$ approaches zero, the choice becomes increasingly ``greedy'' or deterministic, with the agent almost certainly selecting the action with the highest utility. This encourages exploitation of the currently perceived best strategies.
    $$
    \lim_{\tau \to 0} P(a_i) = \begin{cases}
    1 & \text{if } Q(a_i) = \max(Q), \\
    0 & \text{otherwise}.
    \end{cases}
    $$
    \item \textbf{Zero Temperature ($\tau = 0$)}: This implements a strictly greedy strategy, where no exploration occurs.
    $$
    P(a_i) = \mathbb{I}[Q(a_i) > Q(a_j) \ \forall j].
    $$
\end{itemize}
Manipulating $\tau$ allows us to study how different levels of "rationality" or "randomness" in agent decision-making impact the collective outcome.

\subsubsection{Composition Level Dynamics}

The \textbf{Composition Level} $C$ is defined as the proportion of agents (neurons) in a layer currently choosing the `Compose` strategy:
$$
C = \frac{1}{N} \sum_{i=1}^{N} \mathbb{I}[\text{agent } i \text{ chooses Compose}],
$$
where $N$ is the total number of agents (`num\_players`), and $\mathbb{I}[\cdot]$ is the indicator function. The dynamics of this composition level are crucial for understanding whether the system gravitates towards learning disentangled primitives or relying on statistical shortcuts.

A key concept in this simulation is the \textbf{tipping point}. The collective `Compose` strategy only becomes individually advantageous (a stable equilibrium) if its utility exceeds that of the `Detour` strategy. This requires a minimum composition level, given by:
$$
\frac{\kappa \cdot C}{N} \geq c \implies C \geq \frac{c \cdot N}{\kappa}.
$$
In the simulation code, this is reflected by `tipping\_point\_coop\_level = COMPOSE\_COST / BASE\_KAPPA`. This value represents the critical mass of specialized neurons required for the benefits of compositional learning to outweigh the individual risks of specialization. If the system fails to reach or maintain this level, it will, akin to the "Tragedy of the Commons," converge to a state where all agents prefer the `Detour` strategy, leading to a network reliant on brittle statistical shortcuts rather than robust, generalizable features.

\subsubsection{Simulation Modes and Their Ramifications}

The simulation `mode` parameter allows us to directly model the effects of different regularization techniques, particularly classical dropout and our proposed gradient dropout, on the collective learning game:

\begin{itemize}
    \item \textbf{Classical Dropout Mode}: In this mode, with probability $1-p_d$, an agent is stochastically "deactivated" for a given step. If deactivated, its strategy is effectively fixed as `Detour` for that round, as it makes no contribution to the composable features. This simulates the standard dropout mechanism where neurons are effectively removed from the forward pass. From a game-theoretic perspective, this significantly increases the risk for any individual neuron attempting to `Compose`, as the expected number of co-adapting partners is reduced. This intervention powerfully reinforces the non-compositional (`Detour, Detour, \dots, Detour`) Nash Equilibrium, not by promoting `Compose`, but by making collaborative `Compose` strategies even more precarious. Its benefit, as observed in classical dropout, is in forcing neurons to learn features that are useful in isolation, improving generalization against spurious correlations.

    \item \textbf{Gradient Dropout Mode}: This mode directly implements our proposed regularization. While agents are always active in the forward pass (i.e., contributing to the collective output), with probability $1-p$, an agent's strategy update (its "best response" based on the received "utility signal" or gradient) is temporarily suspended. This means the agent effectively ignores the feedback for that specific step and maintains its current strategy choice. This introduces noise into the learning signals themselves, acting as an annealing mechanism in the iterated game. By stochastically nullifying the feedback that would normally push an agent back to the `Detour` strategy, it creates "windows of opportunity" where an agent experimenting with `Compose` might persist in that strategy. If, during these stochastic windows, a critical mass of other agents also happens to explore or maintain the `Compose` strategy, the collective benefit from `Compose` can increase sufficiently to make it a rational choice for all, potentially allowing the system to escape the "Tragedy of the Commons" and find a more globally optimal, compositional equilibrium. This aligns with our empirical findings, where Gradient Dropout, particularly at higher $p$ values (meaning less gradient nullification, but still some annealing), led to more robust models, especially on complex regression tasks.
\end{itemize}

\begin{figure}[h!]
    \centering
    \includegraphics[width=0.7\textwidth]{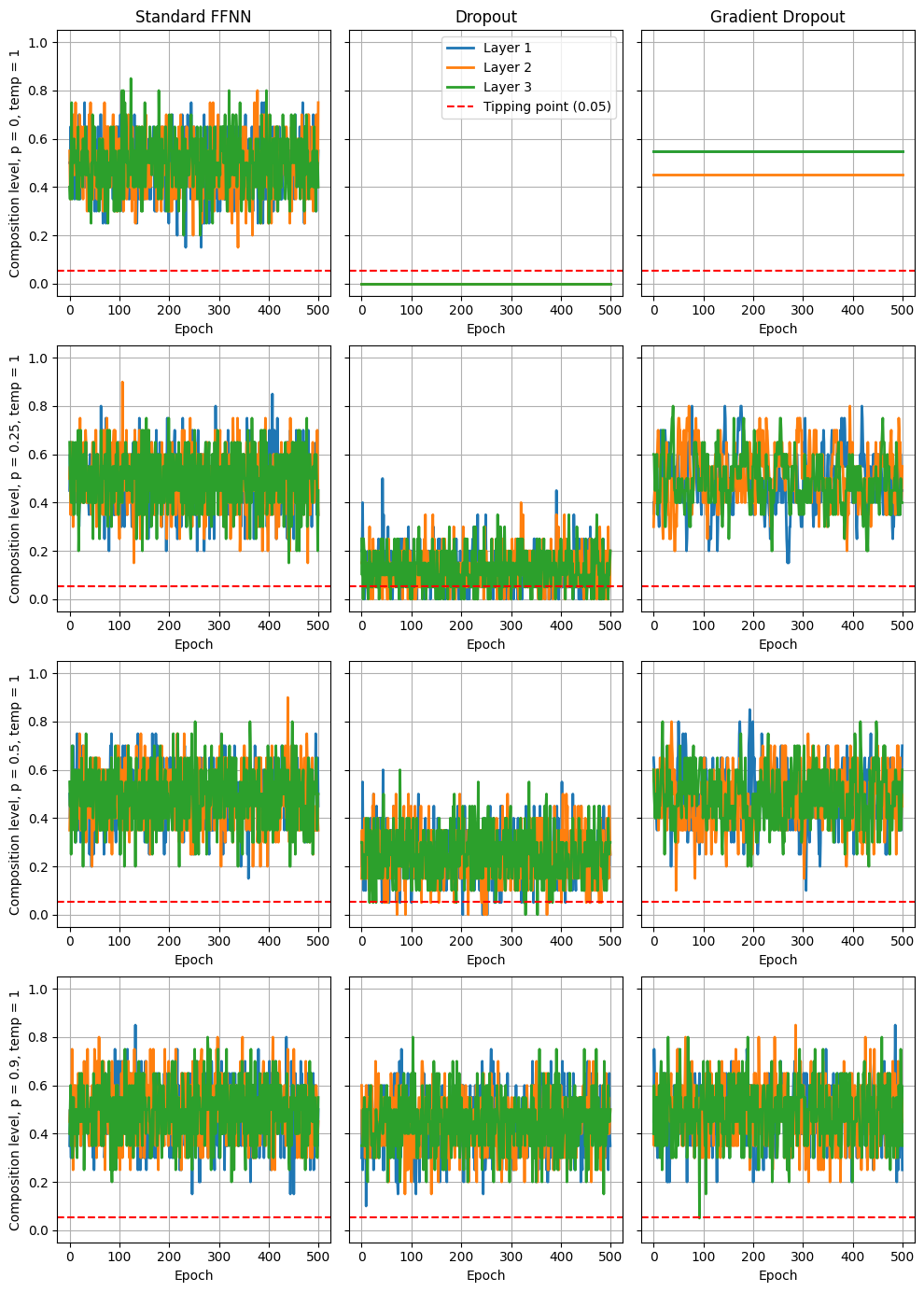}
    \caption{Simulation results showing the evolution of the 'Composition Level' over time under different simulation modes (Classical Dropout vs. Gradient Dropout) and varying temperature $\tau$ or dropout probability $p$. This figure illustrates how different regularization strategies might influence the system's ability to converge on compositional versus shortcut learning equilibria.}
    \label{fig:sim_results}
\end{figure}

The simulation results, as depicted in Figure~\ref{fig:sim_results}, visualize how these different modes influence the evolution of the composition level over training iterations, providing a compelling intuition for the empirical observations on neural network robustness.

By exploring these dynamics in a controlled environment, the simulation reinforces the notion that neural networks can be viewed as complex adaptive systems, where seemingly small interventions in the feedback mechanisms (like gradient dropout) can significantly alter the emergent collective behavior, leading to enhanced robustness or, conversely, to detrimental outcomes depending on the intricate interplay of parameters and problem characteristics. This game-theoretic lens provides a powerful framework for dissecting and designing more robust and interpretable deep learning architectures.

\subsection{Controlled Target Variable Noising}
In addition to the proposed gradient dropout mechanism, we conducted an investigation into the effects of applying controlled noise to the target variable $y$ during the training phase. This approach introduces stochasticity into the objective function (or the utility signal $Y \to R$ in the open game), effectively training the network with an uncertain feedback signal. Our aim was to understand how target noise influences the network's ability to generalize and its robustness when faced with input noise during evaluation. Three distinct strategies for target variable noising were explored:
\begin{enumerate}
    \item \textbf{NoNoise:} The baseline scenario where no artificial noise is added to the target variable $y$. The loss function is computed directly as $L(\hat{y}, y_{\text{target}})$.
    \item \textbf{TDSX:} White noise, sampled from a normal distribution $\mathcal{N}(0, \sigma_y^2)$, is added to $y$. The amplitude of the noise is proportional to the standard deviation of the original target variable of the dataset, $\sigma_y$, multiplied by a factor $X$. Thus, the noisy target $y_{\text{noisy}}$ is given by:
    \begin{equation}
        y_{\text{noisy}} = y_{\text{target}} + X \cdot \sigma_y \cdot \epsilon_{\text{normal}}
    \end{equation}
    where $\epsilon_{\text{normal}} \sim \mathcal{N}(0, 1)$. For the experiments described in the results section, specific values of $X$ were explored (e.g., TDS3, TDS6, TDS9, TDS10, implying $X=3, 6, 9, 10$ respectively).
    \item \textbf{StableNAXBY:} Noise is added to $y$ from a stable distribution. The amplitude is fixed at $0.03 \cdot \sigma_y$. The stable distribution is characterized by its stability parameter $\alpha \in (0, 2]$ and skewness parameter $\beta \in [-1, 1]$. The notation StableNAXBY refers to Dataset N, with specific parameters $\alpha=\text{X}$ and $\beta=\text{Y}$, and a fixed amplitude factor of 0.03. For instance, Stable3A1.25B0F0.03 denotes a strategy applied to Dataset 3 with $\alpha=1.25$, $\beta=0$, and amplitude $0.03 \cdot \sigma_y$. Other tested parameters include $\alpha=1.75$ and $\beta=0$.
    \begin{equation}
        y_{\text{noisy}} = y_{\text{target}} + 0.03 \cdot \sigma_y \cdot \epsilon_{\text{stable}}(\alpha, \beta)
    \end{equation}
\end{enumerate}

\subsection{Datasets}
\DatasetsCount{} publicly available tabular datasets, including both classification and regression tasks with diverse characteristics, were utilized for experimental evaluation. Access URL is available in the Data Availability Statement. A summary of these datasets is provided in Table~\ref{tab:datasets}:

\begin{table}[htbp]
    \caption{Overview of Datasets Used in Experiments.\label{tab:datasets}}
    \centering
    \small
    \begin{tabularx}{\textwidth}{l C C C C}
        \toprule
        \textbf{Dataset} & \textbf{Task} & \textbf{Examples} & \textbf{Features} & \textbf{Source} \\
        \midrule
        wisc\_bc\_data & Classification, 2 classes & 569 & 30 & UCI \\
        car\_evaluation* & Classification, 4 classes & 1728 & 6 & UCI \\
        StudentPerformanceFactors*  & Regression & 6607 & 19 & Kaggle \\
        allhyper & Classification, 2 classes & 3771 & 29 & PMLB \\
        eye\_movements & Classification, 2 classes & 7608 & 23 & HuggingFace \\
        wine\_quality & Regression & 6497 & 11 & HuggingFace \\
        Hill\_Valley\_with\_noise & Classification, 2 classes & 1212 & 100 & PMLB \\
        Hill\_Valley\_without\_noise & Classification, 2 classes & 1212 & 100 & PMLB \\
        294\_satellite\_image & Regression & 6435 & 36 & PMLB \\
        1030\_ERA & Regression & 1000 & 4 & PMLB \\
        \bottomrule
    \end{tabularx}
    \noindent{\footnotesize{* Synthetic dataset.}}
\end{table}


\subsection{Evaluation Metrics}
The performance of the neural networks was evaluated using a set of standard metrics appropriate for each task type:

\begin{itemize}
    \item \textbf{Classification:}
    \begin{itemize}
        \item \textbf{Accuracy:} The proportion of correctly classified instances, offering a fundamental baseline for overall correctness as input noise is introduced.
        \item \textbf{F1-score:} The harmonic mean of precision and recall, important for evaluating performance on potentially imbalanced datasets where accuracy alone can be misleading.
        \item \textbf{ROC AUC (Receiver Operating Characteristic Area Under the Curve):} A measure of the model's intrinsic ability to distinguish between classes, providing a threshold-independent view of robustness against noise-induced shifts in output probabilities.
    \end{itemize}
    \item \textbf{Regression:}
    \begin{itemize}
        \item \textbf{Mean Squared Error (MSE):} MSE was chosen to directly quantify the magnitude of prediction error, enabling a clear assessment of model robustness by tracking how error values change with increasing input noise.
        \item \textbf{Symmetric Mean Absolute Percentage error (SMAPE):} SMAPE was used to provide a scale-independent measure of relative prediction error, making it possible to compare the model's robustness consistently across datasets with different target value ranges.
    \end{itemize}
\end{itemize}
Dependencies of these metrics on the amplitude of uniformly distributed noise added to input features were systematically recorded for all hyperparameter configurations.

\subsection{Reproducibility}
All source code, experimental notebooks, and data supporting this study are publicly available in a GitHub repository: \url{https://github.com/LISA-ITMO/CGT4NN/}. For stable and versioned access to the code corresponding to the findings presented in this paper, we direct readers to the repository's \textit{Releases} page.
\section{Results}

This section describes the experimental results, their interpretation, and the experimental conclusions that can be drawn.

\subsection{Experiments with Input Noise}
Dependencies of quality metrics (Accuracy, F1, ROC AUC for classification; MSE, SMAPE for regression) on the amplitude of uniformly distributed noise added to input features were obtained for various values of the hyperparameter \(p\) (from 0 to 0.99).
\begin{itemize}
    \item On the \textit{wisc\_bc\_data} dataset (№1, classification), an increase in Accuracy at \(p = 0.5\) and \(p = 0.9\) compared to \(p = 0\) (without the proposed method) was observed at various levels of input noise, confirming initial conclusions. At \(p = 0.99\), the best ROC AUC stability was observed.
    \item On the \textit{car\_evaluation} dataset (№2, classification), applying \(p > 0\) generally leads to a decrease in all metrics (F1, Accuracy, ROC AUC) compared to \(p = 0\).
    \item On the \textit{StudentPerformanceFactors} dataset (№3, regression), shown in Figure~\ref{fig:dataset3}, with increasing \(p\), MSE curves become flatter. At \(p = 0.9\), 0.95, 0.99, MSE is significantly lower than at \(p = 0\) for high noise levels. SMAPE is mostly negative.
    \item On the \textit{allhyper} dataset (№4, classification), a complex pattern was observed: F1 and Accuracy show peaks at \(p = 0.01\) for low noise, but at \(p = 0.9\), some robustness is also observed. ROC AUC shows the best results at \(p = 0.01\).
    \item On the \textit{eye\_movements} dataset (№5, classification), applying \(p > 0\) does not provide a clear improvement in metrics compared to \(p = 0\); a decrease is often observed.
    \item On the \textit{wine\_quality} dataset (№6, regression), shown in Figure~\ref{fig:dataset6}, at \(p = 0.9\), the most flattened MSE curve and the lowest MSE values at high noise levels were observed, outperforming \(p = 0\). SMAPE remains low or negative.
    \item On the \textit{Hill\_Valley\_with\_noise} (№7) and \textit{Hill\_Valley\_without\_noise} (№8) datasets (classification), at \(p = 0.01\), some increase in F1 and Accuracy at low and medium noise levels was observed, but ROC AUC is better for \(p = 0\). At high \(p \in \{0.5; 0.9\}\), metrics are usually worse than at \(p = 0\).
\end{itemize}

\subsection{Investigation of Controlled Target Variable Noising}
An investigation was conducted into the influence of three strategies for noising the target variable (\(y\)) during training, combined with various \(p\) values, on the learning process and final robustness to input noise.
\begin{itemize}
    \item \textbf{NoNoise}: Training without adding noise to the target variable. This serves as the baseline scenario.
    \item \textbf{TDSX}: White noise is added to \(y\) with an amplitude proportional to the standard deviation of the original target variable of the dataset, multiplied by a factor \(X\).
    \item \textbf{StableNAXBY}: Noise from a stable distribution is added to \(y\) with an amplitude equal to 0.03 of the standard deviation of the target variable of dataset N, and shape parameters \(\alpha\) and \(\beta\).
\end{itemize}
The goal of this innovation is to investigate how the network's "prior exposure" to noise in target values affects its subsequent robustness to noise in input data and its overall performance.
\begin{itemize}
    \item On the \textit{StudentPerformanceFactors} dataset (№3, regression): The Stable3A1.XBYF0.03 strategy (especially with \(\alpha = 1.25, \beta = 0, p = 0.9\)) shows the best robustness (smallest MSE increase and most stable, albeit negative, SMAPE) to input noise compared to NoNoise and TDS3, as depicted in Figure~\ref{fig:dataset3}.
    \item On the \textit{allhyper} dataset (№4, classification, where the target variable was interpreted as numerical for noising): Stable4A1B1F0.03 with \(p = 0.9\) yields the most stable (though not high) F1, Accuracy, and SMAPE metrics (for ROC AUC, results are less clear) with increasing input noise.
    \item On the \textit{wine\_quality} dataset (№6, regression): The Stable6A1.XBYF0.03 strategy with \(p = 0.9\) (especially \(\alpha = 1.75, \beta = 0\)) demonstrates significant improvement in robustness: MSE grows slower and SMAPE remains more stable (though still low/negative) compared to NoNoise and TDS6, as shown in Figure~\ref{fig:dataset6}.
\end{itemize}

\subsection{Loss Curve Dynamics}
Loss curves during training show that various \(p\) values and target variable noising strategies influence convergence dynamics. High \(p\) values or strong target noising can slow convergence or increase loss function fluctuations in early stages, but in some cases lead to better final performance on noisy data.

\subsection{Emergent Phenomena and Complex System Interpretation}
The ambiguous influence of parameter \(p\) and target noising strategies on various datasets indicates a complex nonlinear dynamic of neuron interaction within the network.
The observed improvement or deterioration in robustness can be interpreted as a result of changes in the emergent behavior of the system (neural network) in response to modifying agent interaction rules (via \(p\)) and altering the "game environment" (via target noising).
In some cases, managing "feedback" for a subset of agents (\(p < 1\)) and training under conditions of target function uncertainty (noising \(y\)) contributes to the formation of more robust collective system behavior strategies, which can be seen as an adaptive emergent property.
This opens perspectives for analyzing neural networks as complex adaptive systems, where transitions to chaotic behavior or self-organization are possible, requiring further investigation using complex systems theory and multi-agent modeling.

\section{Discussion}

The experimental findings reveal the complex interplay between the proposed gradient dropout mechanism, controlled target variable noising, and the inherent characteristics of diverse datasets. This section delves into the interpretation of these results through the lens of compositional game theory (CGT), highlighting the emergent behaviors and implications for neural network robustness.

\subsection{Influence of Gradient Dropout Parameter \(p\)}
Our experiments demonstrate that the hyperparameter \(p\), controlling the probability of a neuron's gradient being included in the backpropagation (thus modulating its "influence" or "voice" in the "co-game" function \(C_G\)), has a highly variable impact across different datasets and tasks.

On datasets like \textit{wisc\_bc\_data} (classification), increasing \(p\) to \(0.5\) and \(0.9\) showed improved Accuracy at various input noise levels, with \(p=0.99\) yielding superior ROC AUC stability. This suggests that a controlled stochastic modulation of agent feedback can indeed enhance robustness, compelling neurons to learn more independent and robust features without fully disabling them on the forward pass, as in classical dropout \citep{Srivastava2014DropoutAS}.

Conversely, on \textit{car\_evaluation}, applying \(p > 0\) consistently led to a degradation of all classification metrics. This indicates that for certain datasets, particularly those with strong inherent structural properties (like the categorical nature of \textit{car\_evaluation}), introducing stochasticity in gradient propagation might impede the learning of critical dependencies, acting as over-regularization or injecting counterproductive noise. The network, as a system of interacting agents, might struggle to converge on effective collective strategies when feedback mechanisms are excessively or inappropriately perturbed.

For regression tasks, such as \textit{StudentPerformanceFactors} (Figure~\ref{fig:dataset3}) and \textit{wine\_quality} (Figure~\ref{fig:dataset6}), higher \(p\) values (e.g., \(0.9\), \(0.95\), \(0.99\)) significantly flattened the MSE curves and reduced MSE at high noise levels compared to \(p=0\). This implies that a reduction in the "reliability" of individual agent feedback, while maintaining overall network connectivity on the forward pass, can foster the development of more resilient "collective strategies" (network weights) that are less susceptible to input perturbations. The negative SMAPE values observed in some regression cases, while indicating poor overall fit for very high noise, are secondary to the primary observation of robustness (flatter MSE curves) as noise increases.

The mixed results across classification and regression tasks, and indeed across different datasets within the same task type (e.g., \textit{allhyper} vs. \textit{eye\_movements}), underscore the complex, non-linear dynamics of agent interaction within the neural network. The optimal level of gradient stochasticity appears to be highly dependent on the intrinsic properties of the data, the nature of the task, and potentially the initial conditions of the optimization landscape.

\subsection{Influence of Controlled Target Variable Noising}
The introduction of controlled noise to the target variable \(y\) during training represents a strategic modification of the "game environment" or the utility function $Y \to R$ that guides the agents' collective learning.

On regression datasets, particularly \textit{StudentPerformanceFactors} (Figure~\ref{fig:dataset3}) and \textit{wine\_quality} (Figure~\ref{fig:dataset6}), the \textit{StableNAXBY} strategy, especially with \(p=0.9\) and specific stable distribution parameters (e.g., \(\alpha=1.25, \beta=0\) or \(\alpha=1.75, \beta=0\)), demonstrated superior noise robustness. This manifested as a slower increase in MSE and more stable (though still low/negative) SMAPE values as input noise increased, outperforming both \textit{NoNoise} and \textit{TDSX} strategies.

This outcome aligns with theoretical expectations: by forcing the network to optimize against a stochastically perturbed target, the agents are implicitly encouraged to seek flatter minima in the loss landscape. As theorized in prior work \citep{chaudhari2019entropy}, flatter minima often correspond to better generalization and increased robustness to various forms of perturbations, including noise in input data. The network, in essence, learns to infer underlying patterns rather than over-fitting to precise target values, making it more resilient to the uncertainty in its objective. The stable distribution noise, with its potential for heavier tails, might provide a more effective form of regularization than Gaussian noise by exposing the network to a broader range of target deviations.

For classification tasks, where the target variable was treated as numerical for noising purposes (e.g., \textit{allhyper}), target noising also showed some benefits, with Stable4A1B1F0.03 and \(p=0.9\) yielding more stable metrics despite not always being the highest. This further suggests that "pre-training" the network with uncertainty in the objective can instill a more robust set of strategies for its agents.

\subsection{Emergent Behaviors in Neural Networks}
Experiments suggest that neural networks, particularly when analyzed through compositional game theory, exhibit characteristics of complex adaptive systems. Performance dependencies on parameters \(p\) and target noising strategies often appear non-linear and sometimes counter-intuitive, hinting at emergent properties from modified neuronal interactions.

The dynamic interplay between controlling gradient flow (via \(p\)) and perturbing the objective function (target noising) seems to foster more robust collective strategies in certain configurations. The neural network, as a system, adapts its weight updates to navigate an environment where both internal feedback and external objectives are uncertain. This adaptation can lead to increased resilience, a valuable emergent property for models operating in noisy real-world scenarios.

This perspective invites further research into neural networks as self-organizing, multi-agent systems, where properties like robustness might be understood as adaptive outcomes of strategic interactions \citep{Cagnetta2024HowDN, Balduzzi2015SemanticsRA}.

Observed shifts in loss convergence dynamics, where higher \(p\) values or stronger target noising occasionally slowed initial convergence but resulted in improved final robustness, support the idea that the system explores a broader, potentially more robust, region of the parameter space.

\subsection{Limitations and Future Directions}
The empirical results, while promising for certain tasks and datasets (especially regression tasks with combined gradient dropout and stable distribution target noising), clearly indicate that there is no universal optimal parameter setting. The efficacy of the proposed methods is highly dependent on the specificity of the dataset and task. This highlights the necessity for adaptive parameter selection mechanisms or more sophisticated meta-learning approaches to dynamically tune these regularization parameters.

Further research should explore the theoretical underpinnings of these emergent behaviors, potentially leveraging advanced tools from complex systems theory to map the parameter space to different regimes of network behavior (e.g., stable, chaotic, self-organizing). Investigating the optimal balance between different types of stochasticity (gradient noise vs. target noise vs. input noise) within the compositional game framework could yield more universally applicable robustness-enhancing techniques. Additionally, extending this framework to more complex neural network architectures and adversarial scenarios would be a natural progression.

\section{Conclusions}


\begin{enumerate}
    \item We proposed and experimentally validated a novel method for enhancing the noise robustness of fully connected neural networks. This method is founded on a game-theoretic interpretation of training as agent interaction, where control is exerted through selective gradient nullification during the error backpropagation phase (gradient dropout) and through the modification of the target function by controlled noising.
    \item Experiments conducted on \datasetscount{} diverse tabular datasets revealed that the proposed method (by varying parameter \(p\)) and the supplementary target variable noising can either improve or, in some instances, degrade robustness to noise and overall accuracy. The outcome is highly dependent on the specific characteristics of the data, the task type, and the chosen hyperparameters. The most promising results concerning robustness enhancement, particularly when combining target noising with stable distributions and a \(p=0.9\) value, were consistently observed in regression tasks.
    \item The study confirmed a dependency of the method's effectiveness on the particular dataset and task, indicating the necessity for adaptive selection of regularization parameters in practical applications.
    \item The obtained results lay the groundwork for future investigations of neural networks as compositions of interacting subsystems (agents). Further development of compositional game theory \citep{cgt, towards-cgt} applied to deep learning holds the potential to foster the creation of more interpretable, adaptive, and robust models capable of exhibiting complex emergent behaviors.
    \item Specific experiments demonstrated that target variable noising, particularly using the StableNAXBY strategy in conjunction with \(p=0.9\) on the regression datasets \textit{StudentPerformanceFactors} (№3, Figure~\ref{fig:dataset3}) and \textit{wine\_quality} (№6, Figure~\ref{fig:dataset6}), led to a notable increase in robustness against input noise, evidenced by more flattened MSE curves and more stable SMAPE values.
\end{enumerate}



\appendix
\section{Appendix}

\begin{figure}[H] 
    \centering
    \includegraphics[width=0.9\textwidth]{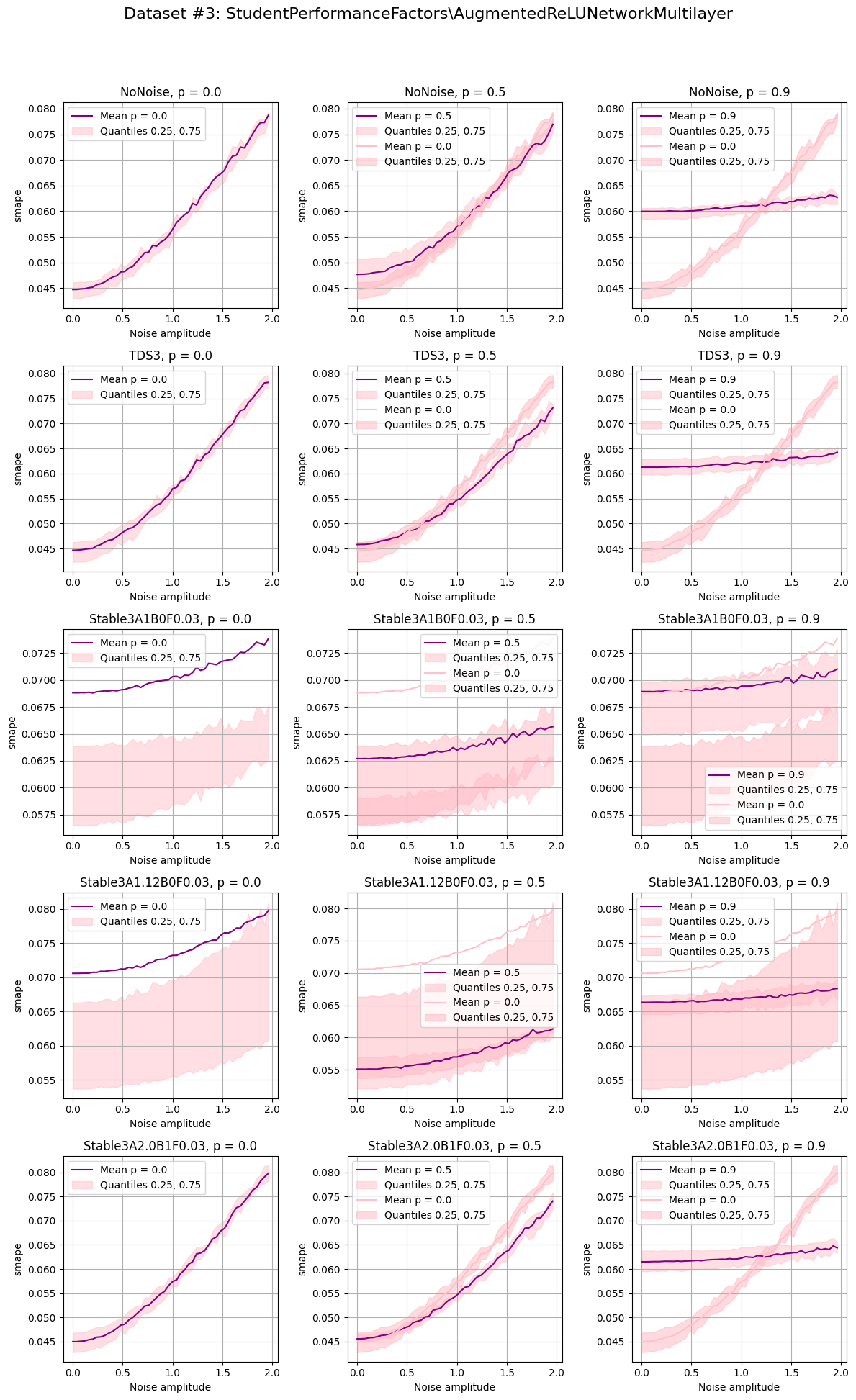}
    \caption{Evaluation curves for the \textit{StudentPerformanceFactors} dataset.}
    \label{fig:dataset3}
\end{figure}

\begin{figure}[H]
    \centering
    \includegraphics[width=0.9\textwidth]{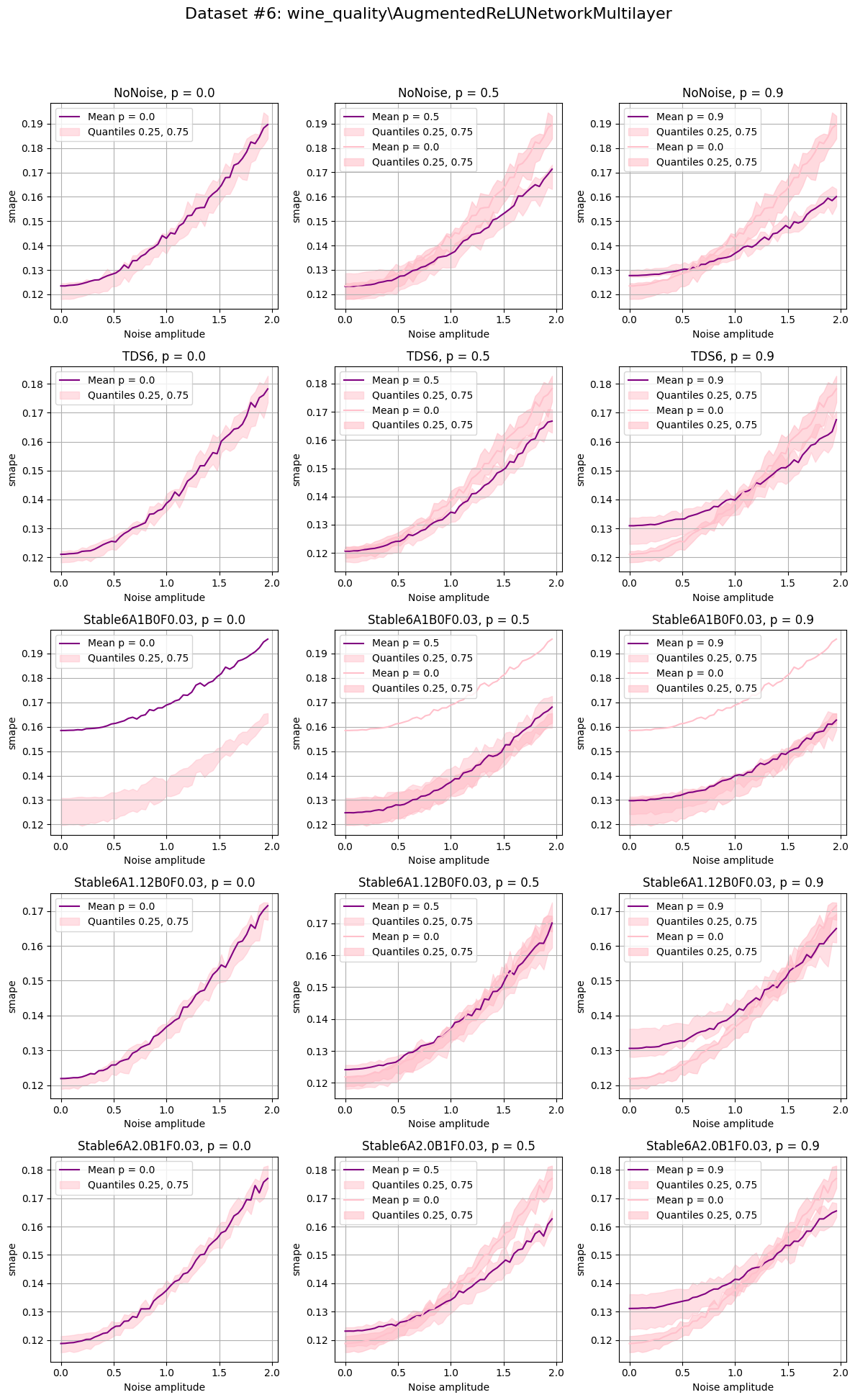}
    \caption{Evaluation curves for the \textit{wine\_quality} dataset.}
    \label{fig:dataset6}
\end{figure}

\begin{figure}[H]
    \centering
    \includegraphics[width=0.9\textwidth]{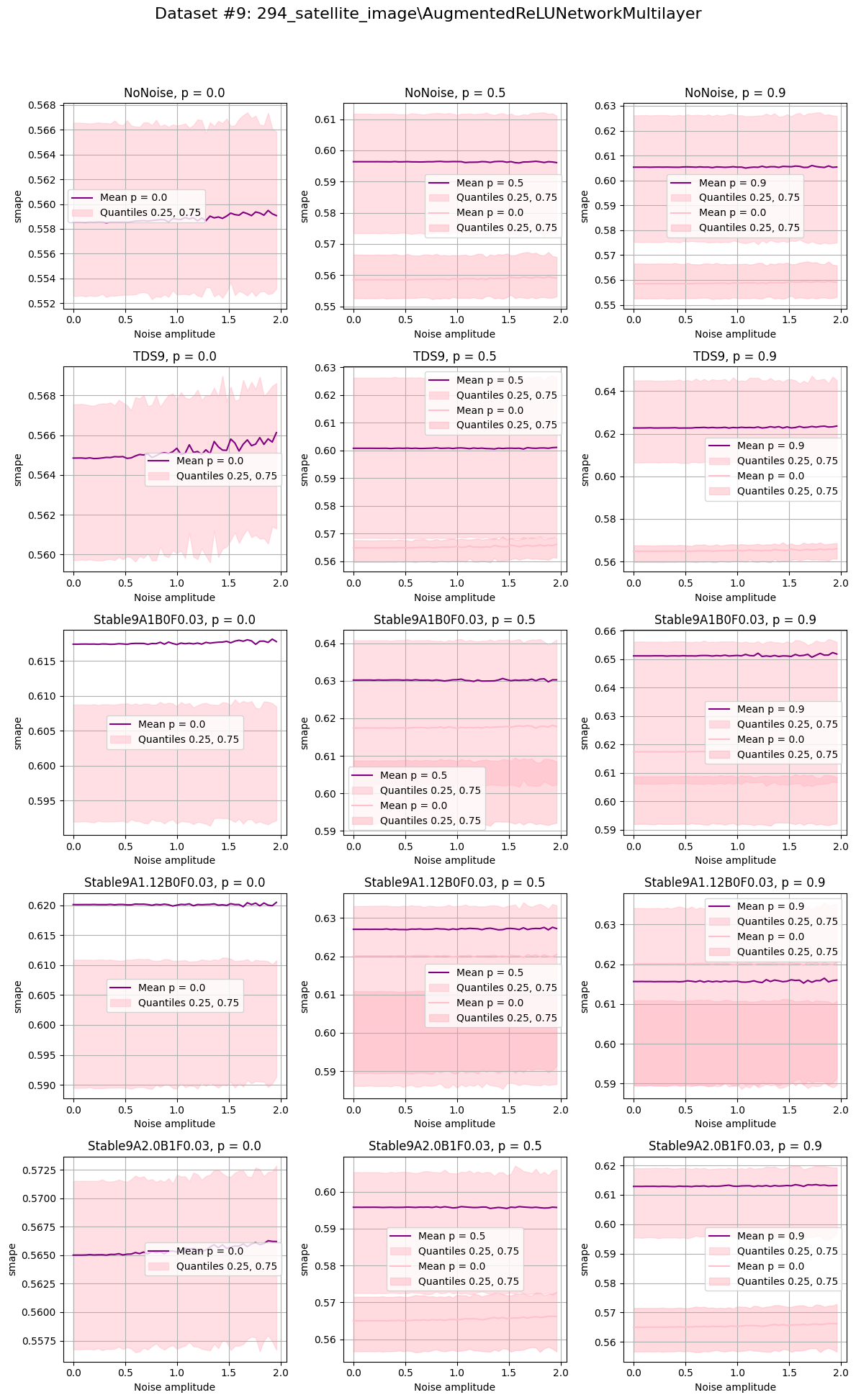}
    \caption{Evaluation curves for the \textit{294\_satellite\_image} dataset.}
    \label{fig:dataset9}
\end{figure}

\begin{figure}[H]
    \centering
    \includegraphics[width=0.9\textwidth]{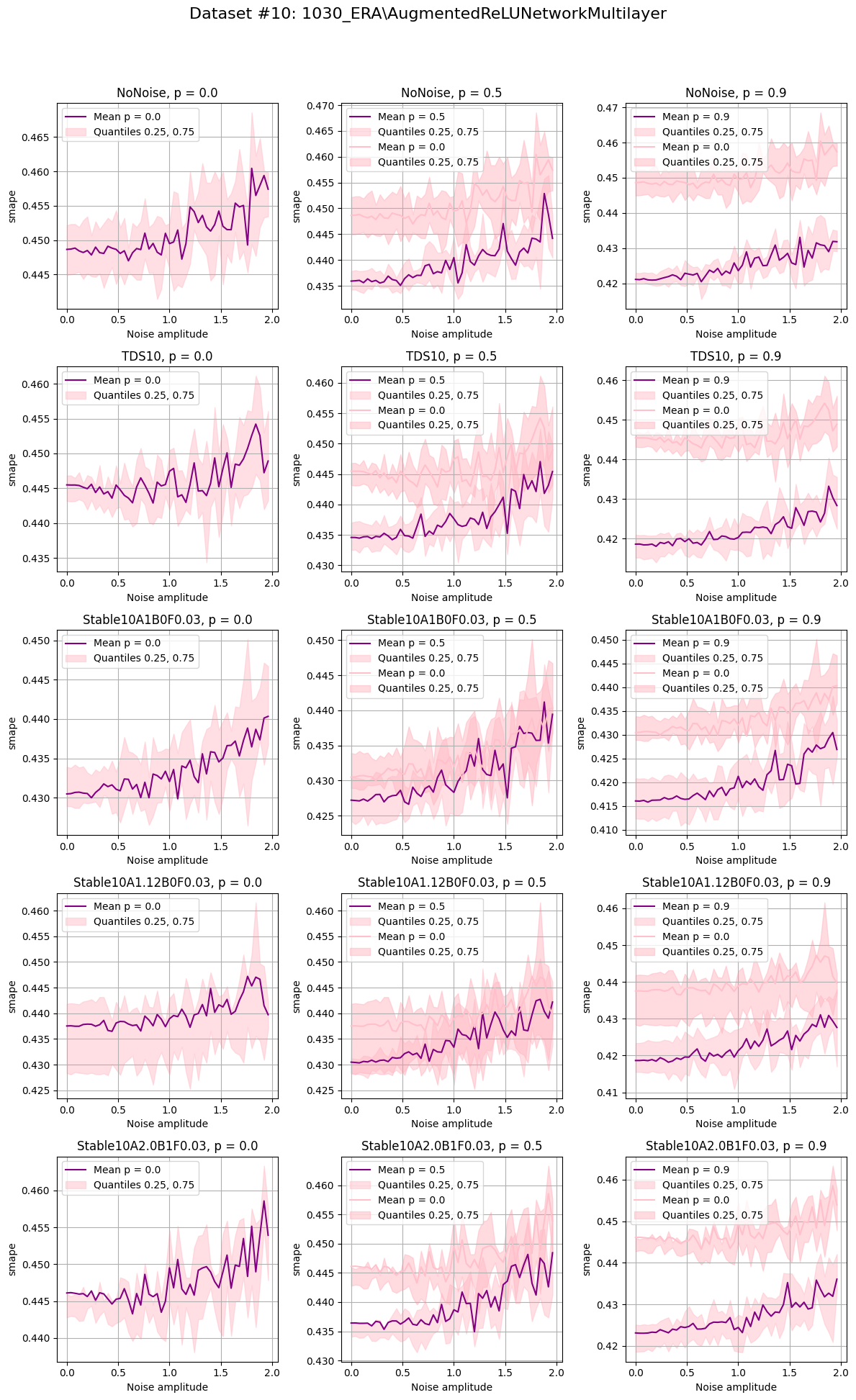}
    \caption{Evaluation curves for the \textit{1030\_ERA} dataset.}
    \label{fig:dataset10}
\end{figure}

\end{document}